\documentclass[10pt,twocolumn,letterpaper]{article}

\usepackage[pagenumbers]{cvpr} 

\definecolor{cvprblue}{rgb}{0.21,0.49,0.74}
\usepackage[pagebackref,breaklinks,colorlinks,allcolors=cvprblue]{hyperref}

\title{Why Does RL Generalize Better Than SFT? A Data-Centric Perspective on VLM Post-Training}

\author{Aojun Lu\\
College of Computer Science,\\
Sichuan University, Chengdu, China\\
{\tt\small aojunlu@stu.scu.edu.cn}
\and
Tao Feng\\
Department of Computer Science and Technology,\\
Tsinghua University, Beijing, China\\
{\tt\small fengtao.hi@gmail.com}
\and
Hangjie Yuan\\
College of Computer Science and Technology,\\
Zhejiang University, Hangzhou, China\\
{\tt\small hj.yuan@zju.edu.cn}
\and
Wei Li\\
College of Computer Science,\\
Sichuan University, Chengdu, China\\
{\tt\small ymjiii98@gmail.com}
\and
Yanan Sun \thanks{Corresponding author}\\
College of Computer Science, Sichuan University, Chengdu, China\\
{\tt\small ysun@scu.edu.cn}
}

\usepackage{multirow}
\usepackage{colortbl}
\usepackage{makecell}

\definecolor{softgreen}{rgb}{0.1, 0.9, 0.1}
\definecolor{softred}{rgb}{0.9, 0.1, 0.1}
\definecolor{lightgray}{rgb}{0.95,0.95,0.95}

\begin{document}
\maketitle

\begin{abstract}
The adaptation of large-scale Vision-Language Models (VLMs) through post-training reveals a pronounced generalization gap: models fine-tuned with Reinforcement Learning (RL) consistently achieve superior out-of-distribution (OOD) performance compared to those trained with Supervised Fine-Tuning (SFT). This paper posits a data-centric explanation for this phenomenon, contending that RL’s generalization advantage arises from an implicit data filtering mechanism that inherently prioritizes medium-difficulty training samples. To test this hypothesis, we systematically evaluate the OOD generalization of SFT models across training datasets of varying difficulty levels. Our results confirm that data difficulty is a critical factor, revealing that training on hard samples significantly degrades OOD performance. Motivated by this finding, we introduce Difficulty-Curated SFT (DC-SFT), a straightforward method that explicitly filters the training set based on sample difficulty. Experiments show that DC-SFT not only substantially enhances OOD generalization over standard SFT, but also surpasses the performance of RL-based training, all while providing greater stability and computational efficiency. This work offers a data-centric account of the OOD generalization gap in VLMs and establishes a more efficient pathway to achieving robust generalization. Code is available at \url{https://github.com/byyx666/DC-SFT}
\end{abstract}

\section{Introduction}

\begin{figure*}[t]
  \centering
  \includegraphics[width=1\linewidth]{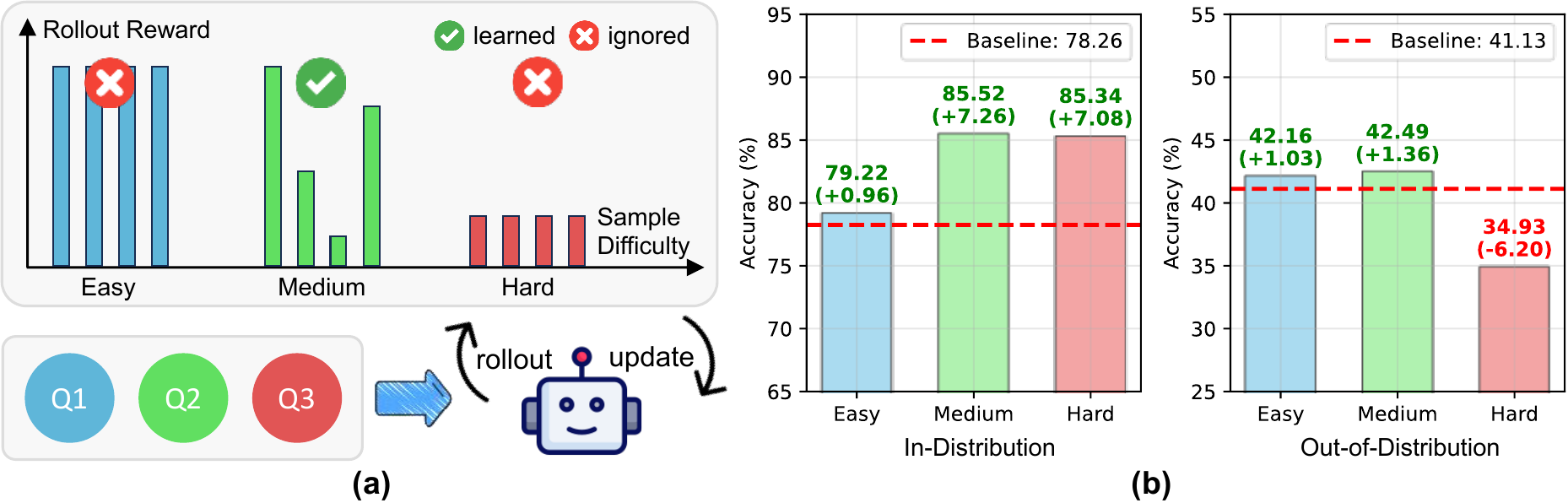}
   \caption{(a) RL implicitly focuses updates on medium-difficulty samples that yield high reward variance. (b) ID and OOD performance after SFT on data subsets of varying difficulty levels.}
   \label{fig:fig1}
\end{figure*}

Adapting large-scale Vision-Language Models (VLMs) to downstream tasks primarily relies on post-training techniques, notably Supervised Fine-Tuning (SFT) and Reinforcement Learning (RL)~\cite{zhang2022opt, achiam2023gpt, guo2025deepseek}. A well-documented finding is that RL-trained models generalize more effectively to out-of-distribution (OOD) data than SFT-trained models, which tend to overfit their in-distribution (ID) training set~\cite{chu2025sft, chen2025synergy, liu2025can, meng2025mm}. This consistent performance gap raises a fundamental question: What underlying mechanism accounts for RL's strong generalization capability?

Prevailing explanations often attribute this advantage to RL's optimization objective, which promotes exploration and learning from reward feedback, thereby enabling generalization beyond the supervised examples used in SFT~\cite{zhang2025reinforcement, shenfeld2025rl}. In this work, we propose a new, data-centric perspective. We posit that the core difference lies in how SFT and RL learn from examples: SFT applies uniform updates across all training data, while RL implicitly prioritizes samples based on their difficulty. As illustrated in Figure~\ref{fig:fig1}a, RL primarily updates the model on medium-difficulty samples that yield high-variance rewards, while effectively ignoring easy (consistently high reward) or hard (consistently low reward) samples~\cite{guo2025deepseek, ahmadian2024back}. This discriminatory updating mechanism leads to our central hypothesis: Does RL's superior generalization arise from its implicit focus on a medium-difficulty data distribution?

To empirically validate this hypothesis, we systematically investigate the influence of training data difficulty on OOD generalization. Specifically, we train SFT models on distinct data subsets curated by difficulty (easy, medium, and hard), and compare their ID and OOD performance. The results, partially shown in Figure~\ref{fig:fig1}b, provide a clear insight: training on hard samples improves ID performance but leads to severe OOD degradation, whereas training on medium-difficulty data achieves a more balanced and robust performance. This suggests that standard SFT is hindered by its uniform treatment of hard samples, while RL succeeds by implicitly concentrating on the critical medium-difficulty subset. Interestingly, training on easy samples also preserves OOD performance, indicating that the optimal data distribution for generalization may be a combination of easy and medium-difficulty samples.

Building on this finding, we propose Difficulty-Curated SFT (DC-SFT), a simple yet effective modification to SFT that explicitly filters out hard data prior to training. Our results demonstrate that this straightforward, data-centric intervention significantly enhances the OOD performance of SFT, surpassing the generalization ability of RL. Critically, DC-SFT achieves these gains with greater computational efficiency and training stability than standard RL pipelines.

The advantages of DC-SFT extend beyond standard OOD benchmarks to complex reasoning tasks. When trained on datasets with explicit reasoning chains, DC-SFT outperforms both standard SFT and RL in enhancing the model's reasoning capabilities. Given its simplicity, stability, and efficiency, DC-SFT emerges as a promising method for developing models with robust generalization and reasoning abilities.

Our contributions are summarized as follows:
\begin{itemize}[leftmargin=*]
\item We propose a novel, data-centric explanation for the generalization gap between RL and SFT, attributing it to RL's role as an implicit data filter that prioritizes medium-difficulty samples.
\item We introduce DC-SFT, a simple yet efficient method that explicitly filters out hard data prior to training. DC-SFT surpasses RL's OOD generalization while being more stable and computationally efficient.
\item We demonstrate that DC-SFT delivers superior performance on complex reasoning tasks, underscoring its potential as a powerful method for fostering generalized reasoning abilities.
\end{itemize}

\section{Related Works}

\textbf{Post-training of Foundation Models.}
Post-training is a critical phase that adapts pre-trained foundation models for effective application in downstream tasks~\cite{zhang2022opt, achiam2023gpt, guo2025deepseek}. SFT and RL are two primary paradigms used in this stage~\cite{chu2025sft}. SFT~\cite{radford2018improving, weifinetuned, zhou2023lima} involves training models on curated datasets of instruction-response pairs, equipping them to comprehend human instructions and generate appropriate responses. Prior research has demonstrated that SFT enables models to tackle a wide spectrum of vision and multimodal tasks, ranging from simple image classification~\cite{jiang2024supervised} to complex vision-language reasoning~\cite{shao2024visual}. In contrast, RL~\cite{ziegler2019fine, zhou2024archer, zhai2024fine} trains a model to maximize reward signals, which are typically derived from human preferences or rule-based metrics. While RL has traditionally been employed to better align model outputs with human preferences~\cite{ziegler2019fine, ouyang2022training}, recent research has expanded its focus to enhancing performance on specific, objective-driven tasks~\cite{guo2025deepseek, shen2025vlm}. For instance, Visual-RFT~\cite{liu2025visual} enhances VLMs by leveraging RL with rule-based visual rewards, resulting in models that exhibit superior performance on various visual tasks compared to those trained with traditional SFT.

\textbf{Generalization of SFT and RL.}
Generalization is a critical attribute for VLMs, given the inherent distribution shifts in real-world applications~\cite{zhang2024out, mayilvahanan2024search}. Existing studies indicate that LLMs and VLMs tend to overfit more on simpler, knowledge-intensive tasks, while demonstrating stronger generalization on more complex, reasoning-intensive ones~\cite{wang2024generalization, qi2024quantifying, tong2024metamorph}. Beyond these findings, other research highlights the significant influence of post-training paradigms on model generalization~\cite{chu2025sft, liu2025can, chen2025synergy}. For example, a systematic comparison across both textual and visual tasks shows that RL is more effective at learning generalizable knowledge, whereas SFT is prone to memorizing the training data~\cite{chu2025sft}. Building on this finding, our work aims to investigate the underlying reasons for RL's enhanced generalization from a data-centric perspective.

\textbf{Data Curation for SFT.}
Constructing high-quality instruction datasets through data synthesis has become a cornerstone for improving model performance via SFT~\cite{xin2024deepseek, guo2025deepseek}. Two prominent techniques in this domain are rejection sampling and model distillation. Rejection sampling methods create high-quality data by prompting the target model to generate multiple responses for a given instruction and then selecting the optimal one~\cite{roziere2023code, wang2023self}. Model distillation, conversely, bypasses this iterative process by employing a well-trained teacher model to directly produce high-quality responses~\cite{chen2024sharegpt4v, liu2023visual}. In both cases, the synthesized instruction–response pairs constitute refined training data for SFT. Notably, model distillation has proven effective for SFT, demonstrating performance that can match or even surpass that of RL in certain scenarios~\cite{guo2025deepseek}. In this study, we demonstrate that focusing on easy and medium-difficulty samples has the potential to further improve existing data synthesis in terms of generalization.

\textbf{Curriculum Learning.}
Curriculum learning seeks to enhance model training by gradually introducing examples in a meaningful order, typically progressing from easier to more difficult instances~\cite{bengio2009curriculum}. Within the field of VLMs, curriculum learning has shown considerable promise for improving model performance and generalization~\cite{sheng2025curricuvlm}. While curriculum learning primarily focuses on the sequencing of training data, our study instead centers on enhancing generalization by filtering data.

\textbf{Test Time Scaling.}
Recent empirical research has shown that encouraging models to extend
the responses before producing a final answer can significantly improve model accuracy~\cite{wei2022chain, yao2023tree, team2025kimi}. This observation highlights the ``scaling laws'' for test-time computation: allocating more computational resources during inference can lead to better performance~\cite{jaech2024openai, guo2025deepseek}. Early studies in this area primarily use prompting strategies to elicit intermediate reasoning steps before the final answer~\cite{wei2022chain, yao2023tree}. More recently, the field has shifted toward using RL to scale test-time computation, which demonstrates high performance and strong generalization on both textual and visual domains~\cite{guo2025deepseek, team2025kimi}. An important nuance arises from recent findings: for relatively small models, SFT combined with model distillation can outperform RL in reasoning abilities~\cite{guo2025deepseek}. This suggests a promising direction for SFT-based test-time scaling, though the comparatively limited generalization of SFT relative to RL may pose a challenge. Our work proposes a solution to bridge this generalization gap between SFT and RL, thereby enabling more effective use of SFT for scaling test-time computation.

\section{Preliminaries}

Post-training refers to the process of updating the parameters $\theta$ of a pre-trained foundation model $\pi_\theta$ to enhance its performance on specific downstream tasks or to better align its behavior with human preferences. This is commonly accomplished through SFT or RL, applied to a dataset $\mathcal{D}$.

\textbf{Supervised Fine-Tuning (SFT).} Given a labeled dataset $\mathcal{D} = \{(x_i, y_i)\}_{i=1}^N$, where each $x_i$ is a prompt and $y_i$ is the corresponding desired response, SFT aims to maximize the conditional likelihood of generating $y_i$ given $x_i$. This is achieved by minimizing the negative log-likelihood loss:
\begin{equation}
    \mathcal{L}_{\text{SFT}}(\theta) = -\mathbb{E}_{(x,y) \sim \mathcal{D}} \sum_{t=1}^{|y|} \log \pi_\theta(y_t \mid x, y_{<t}),
\end{equation}
where $|y|$ denotes the length of the response, $y_t$ is the $t$-th token, and $y_{<t}$ represents all tokens preceding $y_t$.

\textbf{Reinforcement Learning (RL).} In RL-based post-training, the model is treated as a policy, and the token generation process is formulated as a Markov Decision Process (MDP). Specifically, each input prompt $x$ defines an initial state, and generating each token corresponds to taking an action. The model transitions between states by sequentially sampling tokens until a termination condition is satisfied (e.g., an end-of-sequence token)~\cite{fu2025srft}. A reward function $R$ is designed to evaluate the quality of the generated responses, often based on human preferences~\cite{ziegler2019fine, ouyang2022training} or task-specific metrics~\cite{guo2025deepseek, shen2025vlm}. The goal is to maximize the expected cumulative reward:
\begin{equation}
    J_{\text{RL}}(\theta) = \mathbb{E}_{x \sim \mathcal{D}, \, y \sim \pi_\theta(\cdot \mid x)} [ r(x, y) ].
\end{equation}
To stabilize training and prevent excessive deviation from the initial reference model, a penalty term is often incorporated, leading to the following objective:
\begin{equation}
    \mathcal{L}_{\text{RL}}(\theta) = -J_{\text{RL}}(\theta) + \beta \mathbb{D}_{\text{KL}}( \pi_\theta(\cdot \mid x) \parallel \pi_{\theta_{\text{ref}}}(\cdot \mid x) ),
\end{equation}
where $\beta > 0$ is a tuning parameter that controls the strength of the Kullback–Leibler (KL) divergence penalty.

Our study focuses on Group Relative Policy Optimization (GRPO)~\cite{shao2024deepseekmath} as a representative RL algorithm, given its proven effectiveness and growing popularity in recent research. The GRPO objective function is defined as follows:
\begin{equation}
\begin{aligned}
    &\mathcal{J}_{\text{GRPO}}(\theta) = \frac{1}{G} \sum_{k=1}^{G} \frac{1}{|y^k|} \sum_{t=1}^{|y^k|} I,
    \\
    &I = \min\{ r^k_{t}(\theta) \cdot {A}^k,\ \text{clip}(r^k_{t}(\theta), 1-\epsilon, 1+\epsilon) \cdot {A}^k \},
    \\
    &r^k_{t}(\theta) = \frac{\pi_\theta(y^k_{t} \mid x, y^k_{<t})}{\pi_{\theta_{\text{old}}}(y^k_{t} \mid x, y^k_{<t})},
\end{aligned}
\end{equation}
where $G$ denotes the number of responses sampled per prompt, $y_k$ denotes the $k$-th response for the prompt $x$. ${A}^k$ represents the normalized advantage computed from the reward signals within the group, specifically:
\begin{equation}
\label{eq:advantage}
A^{k}=\frac{r(x, y^{k})-\operatorname{mean}(\{r(x, y^{k}) \mid k=1,2, \ldots, G\})}{\operatorname{std}(\{r(x, y^{k}) \mid k=1,2, \ldots, G\}) + \delta},
\end{equation}
where $\delta$ is a small constant added for numerical stability.

\begin{figure*}[t]
  \centering
  \includegraphics[width=1\linewidth]{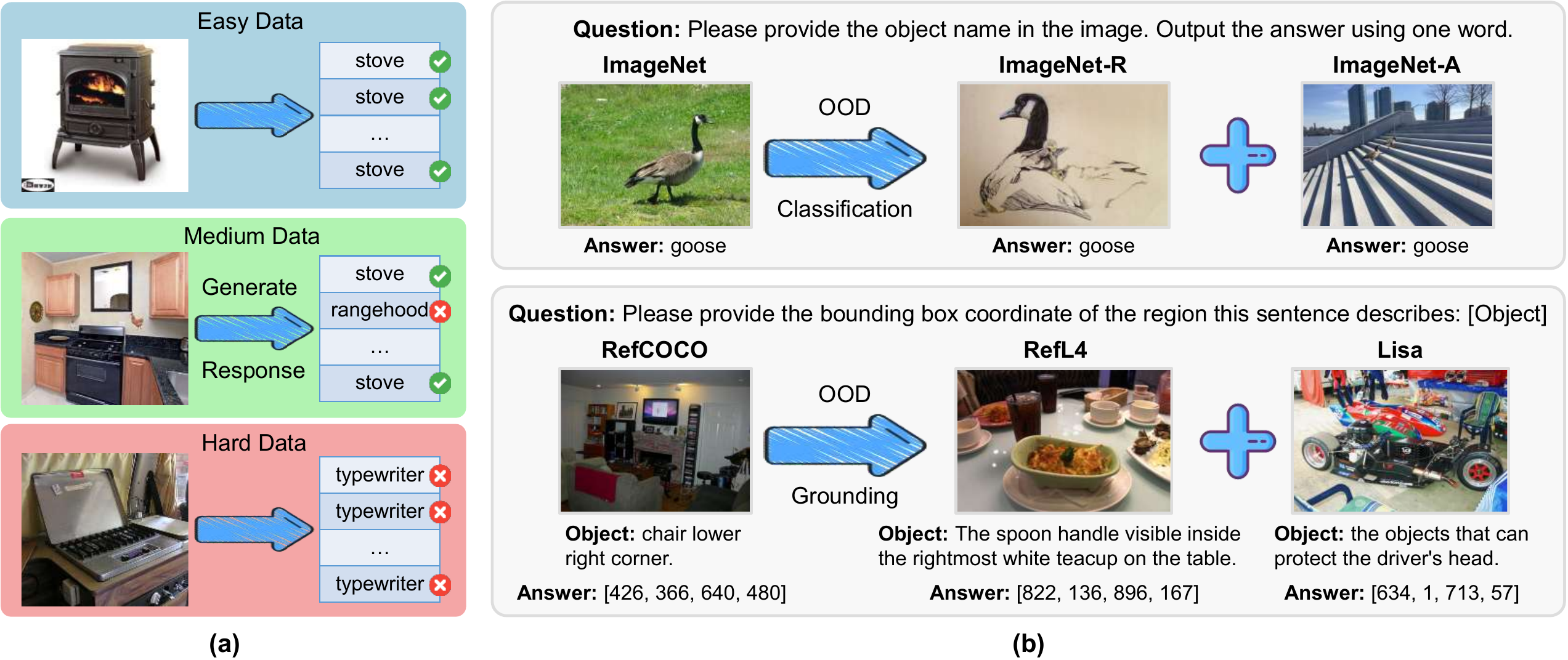}
   \caption{(a) Illustrative examples of the data difficulty taxonomy. (b) Illustrative examples of generalization evaluation benchmarks for image classification (top) and visual grounding (bottom).}
   \label{fig:benchmark}
\end{figure*}

\section{Data-Centric Analysis of Generalization}

In this section, we first outline our core data-centric hypothesis, arguing that RL's generalization advantage primarily stems from implicit data selection rather than from algorithmic properties. We then present experiments examining how training data of varying difficulty influences generalization, providing empirical support for this hypothesis.

\subsection{The Data-Centric Hypothesis of Generalization}
\label{sec:analysis}

Although prior research has explored RL’s generalization capabilities from multiple perspectives, we adopt a data-centric perspective to explain why RL often generalizes better than SFT and how this understanding can guide further improvements. Specifically, we hypothesize that RL's generalization advantage relative to SFT can be attributed to an implicit data filtering mechanism that concentrates learning on medium-difficulty examples.

\textbf{A Taxonomy of Data Difficulty.}
To illustrate this mechanism, we categorize training samples based on the model's perceived difficulty. As shown in Figure~\ref{fig:benchmark}a, for a given prompt $x$, the VLM generates $G$ distinct responses (in this study, $G=8$). We then classify each instance as follows:
\begin{itemize}
\item \textit{Easy}, if all $G$ responses are correct;
\item \textit{Hard}, if all $G$ responses are incorrect;
\item \textit{Medium-difficulty}, if the responses contain a mixture of correct and incorrect answers.
\end{itemize}

\textbf{Inherent Data Filter within RL.}
Under this taxonomy, the RL update process inherently filters data based on difficulty. For both easy instances (where all responses are correct and receive maximal reward) and hard instances (where all responses are incorrect and receive minimal reward), the rewards are uniform across all generated responses. This reward uniformity leads to a relative advantage $A^k = 0$ for every response associated with such instances, as defined by Eq.~\ref{eq:advantage}. As a result, these instances contribute negligible gradient updates during training. In contrast, medium-difficulty instances yield diverse rewards, producing meaningful advantage estimates and gradients. Consequently, the RL-based training is dominated by these medium-difficulty examples, effectively filtering out easy and hard instances.

\begin{table*}[htbp]
\centering
\setlength{\tabcolsep}{1.5mm}
\caption{ID and OOD (gray background) performance (\%) after SFT on data subsets of varying difficulty levels. The baseline denotes the performance of the initialized model without subsequent fine-tuning. Performance changes exceeding 1\% are highlighted in color.}

\label{tab:difficulty}
    \begin{tabular}{ccc>{\columncolor{lightgray}}c>{\columncolor{lightgray}}cc>{\columncolor{lightgray}}c>{\columncolor{lightgray}}c}
    \toprule
    \multirow{2}{*}{Model} & \multirow{2}{*}{SFT Data} & \multicolumn{3}{c}{Image Classification} & \multicolumn{3}{c}{Visual Grounding} \\
    \cmidrule(lr){3-5} \cmidrule(lr){6-8}
    & & ImageNet & ImageNet-R & ImageNet-A & RefCOCO & Ref-L4 & Lisa \\
    \midrule
    \multirow{4}{*}{\makecell[c]{Qwen2.5-\\VL-3B}} 
    & Baseline & 76.62 & 60.51 & 31.94 & 88.44 & 73.45 & 64.84 \\
    & Easy & 76.20 (-0.42) & 60.00 (-0.51) & 33.30 (\textcolor{softgreen}{+1.36}) & 89.17 (+0.73) & 74.42 (+0.97) & 64.78 (-0.06) \\
    & Medium & 80.36 (\textcolor{softgreen}{+3.74}) & 59.13 (\textcolor{softred}{-1.38}) & 32.94 (\textcolor{softgreen}{+1.00}) & 89.45 (\textcolor{softgreen}{+1.01}) & 73.61 (+0.16) & 64.72 (-0.12) \\
    & Hard & 81.16 (\textcolor{softgreen}{+4.54}) & 51.17 (\textcolor{softred}{-9.34}) & 26.90 (\textcolor{softred}{-5.04}) & 88.25 (-0.19) & 72.11 (\textcolor{softred}{-1.34}) & 61.34 (\textcolor{softred}{-3.50}) \\
    \midrule
    \multirow{4}{*}{\makecell[c]{Qwen2.5-\\VL-7B}}
    & Baseline & 78.26 & 57.32 & 41.13 & 85.04 & 69.25 & 68.70 \\
    & Easy & 79.22 (+0.96) & 59.37 (\textcolor{softgreen}{+2.05}) & 42.16 (\textcolor{softgreen}{+1.03}) & 86.02 (+0.98) & 70.73 (\textcolor{softgreen}{+1.48}) & 70.27 (\textcolor{softgreen}{+1.57}) \\
    & Medium & 85.52 (\textcolor{softgreen}{+7.26}) & 58.67 (\textcolor{softgreen}{+1.35}) & 42.49 (\textcolor{softgreen}{+1.36}) & 87.05 (\textcolor{softgreen}{+2.01}) & 71.32 (\textcolor{softgreen}{+2.07}) & 69.84 (\textcolor{softgreen}{+1.14}) \\
    & Hard & 85.34 (\textcolor{softgreen}{+7.08}) & 43.25 (\textcolor{softred}{-14.07}) & 34.93 (\textcolor{softred}{-6.20}) & 86.54 (\textcolor{softgreen}{+1.50}) & 69.68 (+0.43) & 67.73 (-0.97) \\
    \bottomrule
    \end{tabular}
\end{table*}

\subsection{Experimental Setup}
\label{sec:settings}

\textbf{Evaluation Tasks and Benchmarks.} To systematically assess generalization, we evaluate the models on two representative vision-language tasks: image classification and visual grounding. Specifically, we train each model on a task-specific dataset and then evaluate its performance on both ID and OOD test sets. Illustrative examples of these benchmarks are shown in Figure~\ref{fig:benchmark}b.

\textit{Image Classification Benchmark.} We construct a training subset from ImageNet-1K~\cite{ridnik2021imagenet} by randomly selecting 100 single-word object categories, each containing 100 images. ID evaluation is performed on the official ImageNet-1K test set corresponding to these 100 classes. For OOD evaluation, we employ ImageNet-R (artistic renditions)~\cite{hendrycks2021many} and ImageNet-A (natural adversarial examples)~\cite{hendrycks2021natural}, both of which exhibit significant distribution shifts from ImageNet-1K.

\textit{Visual Grounding Benchmark.} We build a training set by sampling 10,000 instances from the RefCOCO training set~\cite{mao2016generation,yu2016modeling}, ensuring each image appears only once. ID evaluation is conducted on the RefCOCO validation set. For OOD evaluation, we use Ref-L4~\cite{chen2025revisiting} and Lisa~\cite{lai2024lisa}, which feature referring expressions with linguistic and visual properties distinct from the training data. Specifically, we use a subset of the Ref-L4 validation set that excludes all COCO images to ensure a sufficient domain gap. For Lisa, we adopt a version with bounding box annotations (rather than segmentation masks) to maintain consistency, following VLM-R1~\cite{shen2025vlm}.

\textbf{Training Settings.} We utilize Qwen2.5-VL-3B and Qwen2.5-VL-7B~\cite{bai2025qwen2} as the foundational VLMs. All models are first initialized with one epoch of full fine-tuning on a randomly selected subset of 400 samples, which is excluded from all subsequent training. For the subsequent post-training phase, we employ parameter-efficient LoRA fine-tuning~\cite{hu2022lora} with a rank of 32 and alpha of 64. Across both stages, we use the AdamW optimizer with an initial learning rate of $1 \times 10^{-5}$, a batch size of 16, and a cosine learning rate scheduler with a warm-up ratio of 0.1.

We employ the initialized model to generate eight distinct responses ($G=8$) for every training instance using a temperature of 0.9 and a top-p of 1.0. Based on the correctness of these responses, we partition the training data into the easy, medium-difficulty, and hard subsets as defined in Section~\ref{sec:analysis}. To isolate the impact of data difficulty, we conduct SFT separately on each subset. Given that the number of instances varies significantly among different subsets, we uniformly subsample each category to align with the size of the smallest subset to prevent potential confounding effects. All models are trained for 100 steps.

\textbf{Evaluation Settings.} During evaluation, we adopt greedy decoding with a temperature of 0, ensuring deterministic outputs. For \textit{image classification}, a response is considered correct only if the generated label exactly matches the ground-truth label (case-insensitive). For \textit{visual grounding}, a prediction is deemed correct if the Intersection over Union (IoU) between the predicted and ground-truth bounding boxes is 0.5 or greater. In all experiments, we report the accuracy on each test dataset.

\textbf{Implementation Details} For all experiments, we utilize the ms-swift~\cite{zhao2024swiftascalablelightweightinfrastructure} framework. The maximum pixel count is set to 802,816 (corresponding to an image size of 896 $\times$ 896) for all models. To adhere to this maximum pixel constraint, images are resized to ensure they do not exceed 896 $\times$ 896 pixels, and the labels for visual grounding are adjusted accordingly.

\subsection{Performance Results} 

Our experimental results reveal a consistent and pronounced effect of training data difficulty on model performance. As summarized in Table~\ref{tab:difficulty}, models trained exclusively on hard samples achieve strong ID performance but exhibit substantial degradation on OOD benchmarks. For instance, fine-tuning Qwen2.5-VL-7B on the hard subset of ImageNet yields a 7.08\% improvement in ID accuracy over the baseline, yet incurs a severe 14.07\% performance drop on ImageNet-R. In contrast, models trained on medium-difficulty data demonstrate a more balanced performance profile, achieving strong ID improvements while preserving or slightly enhancing OOD performance relative to the baseline. Interestingly, models trained on easy samples maintain stable OOD performance, though they consistently underperform models trained on medium-difficulty samples on ID tasks. Collectively, these findings support our hypothesis that RL's generalization advantage can be attributed, at least partially, to its implicit focus on medium-difficulty examples. Standard SFT, by contrast, treats all samples uniformly and is consequently influenced by hard examples that hinder OOD generalization.

\begin{table*}[htbp]
\centering
\setlength{\tabcolsep}{1mm}
\caption{ID and OOD (gray background) performance (\%) of different post-training paradigms on image classification and visual grounding. \textbf{Bolded} indicates the best, and \underline{underline} indicates the second-best. Performance improvement is calculated relative to standard SFT.}
\label{tab:dcsft}
\resizebox{\linewidth}{!}{
\begin{tabular}{ccc>{\columncolor{lightgray}}c>{\columncolor{lightgray}}cc>{\columncolor{lightgray}}c>{\columncolor{lightgray}}cc}
\toprule
\multirow{2}{*}{Model} & \multirow{2}{*}{Paradigms} & \multicolumn{3}{c}{Image Classification} & \multicolumn{3}{c}{Visual Grounding} & \multirow{2}{*}{OOD Avg.} \\
\cmidrule(lr){3-5} \cmidrule(lr){6-8}
& & ImageNet & ImageNet-R & ImageNet-A & RefCOCO & Ref-L4 & Lisa \\
\midrule
\multirow{4}{*}{\makecell[c]{Qwen2.5-\\VL-3B}}
& SFT & \textbf{91.88} & 54.22 & 34.63 & 90.45 & 73.26 & 63.87 & 56.50 \\
& GRPO & 89.30 (-2.58) & 57.66 (+3.44) & 35.16 (+0.53) & \textbf{90.62} (+0.17) & \textbf{74.54} (+1.28) & \textbf{65.86} (+1.99) & 58.31 (+1.81) \\
& SFT-M & \underline{90.18} (-1.70) & \underline{60.14} (+5.92) & \textbf{37.61} (+2.98) & \underline{90.59} (+0.14) & \underline{74.05} (+0.79) & 64.29 (+0.42) & \underline{59.02} (+2.52) \\
& SFT-EM & 89.38 (-2.50) & \textbf{61.74} (+7.52) & \underline{37.25} (+2.62) & 90.32 (-0.13) & 73.85 (+0.59) & \underline{64.90} (+1.03) & \textbf{59.44} (+2.94) \\
\midrule
\multirow{4}{*}{\makecell[c]{Qwen2.5-\\VL-7B}}
& SFT & \textbf{93.70} & 49.58 & 39.14 & \underline{89.99} & 72.15 & 69.60 & 57.62 \\
& GRPO & 90.80 (-2.90) & 51.38 (+1.80) & \underline{43.65} (+4.51) & 89.91 (-0.08) & 71.60 (-0.55) & \textbf{71.29} (+1.69) & 59.48 (+1.86) \\
& SFT-M & \underline{92.22} (-1.48) & \underline{55.79} (+6.21) & 42.49 (+3.35) & \textbf{90.17} (+0.18) & \underline{73.20} (+1.05) & 70.14 (+0.54) & \underline{60.41} (+2.79) \\
& SFT-EM & 91.44 (-2.26) & \textbf{55.90} (+6.32) & \textbf{44.57} (+5.43) & 89.81 (-0.18) & \textbf{73.40} (+1.25) & \underline{70.51} (+0.91) & \textbf{62.10} (+4.48) \\
\bottomrule
\end{tabular}
}
\end{table*}

\section{DC-SFT: Enhancing SFT's Generalization}

In this section, we introduce Difficulty-Curated Supervised Fine-Tuning (DC-SFT), an approach that fine-tunes the models on data subsets curated according to difficulty levels. We propose two variants of DC-SFT: SFT-M, which mimics the implicit data filtering mechanism of RL by training exclusively on medium-difficulty instances; and SFT-EM, which incorporates both easy and medium-difficulty examples to enhance generalization, as evidenced by the benefits observed in earlier experiments.

\subsection{Evaluation Settings} 

We compare DC-SFT against standard SFT and GRPO, both of which are trained on the full dataset. For a fair comparison, GRPO uses the same generation settings as DC-SFT's data pre-filtering, sampling eight responses per prompt with a temperature of 0.9 and top-p of 1.0. To maintain a consistent number of learning steps, we use batch sizes of 16 for SFT and 128 for GRPO. For GRPO, we apply a KL divergence penalty of 0.04 to regularize the model. All models are trained for 600 steps, which is equivalent to one epoch on the full dataset, with all other hyperparameters matching those described in Section~\ref{sec:settings}.

\subsection{Performance Results}

The results in Table~\ref{tab:dcsft} align with the known trade-off~\cite{chu2025sft}: standard SFT excels at ID performance, while RL (GRPO) typically shows better OOD generalization. However, our DC-SFT variants, which explicitly filter the training data, markedly outperform both baseline methods on average OOD metrics while showing high ID performance.

Crucially, SFT-M, which is trained only on medium-difficulty data to mimic RL's implicit filtering, matches or exceeds the OOD performance of the RL (GRPO). Specifically, SFT-M shows an average OOD accuracy gain of 2.79\% over SFT and 0.93\% over GRPO for the Qwen2.5-VL-7B model. This directly supports our central hypothesis: the generalization advantage attributed to RL is not unique to its algorithm but stems from its implicit focus on medium-difficulty samples.

Moreover, the SFT-EM variant, which removes only the hard samples and trains on both easy and medium-difficulty data, achieves the strongest OOD performance. It delivers substantial improvements over standard SFT (up to 4.48\% for the 7B model) and significantly outperforms GRPO (by 2.62\% for the 7B model). This result demonstrates that while RL's focus on medium-difficulty data is beneficial, the primary cause of standard SFT's poor generalization is its uniform inclusion of hard samples. By simply removing this detrimental data subset, SFT-EM establishes a new, more effective baseline for OOD generalization. 

\subsection{Training Stability Analysis}

\begin{figure}[ht]
  \centering
  \includegraphics[width=1\linewidth]{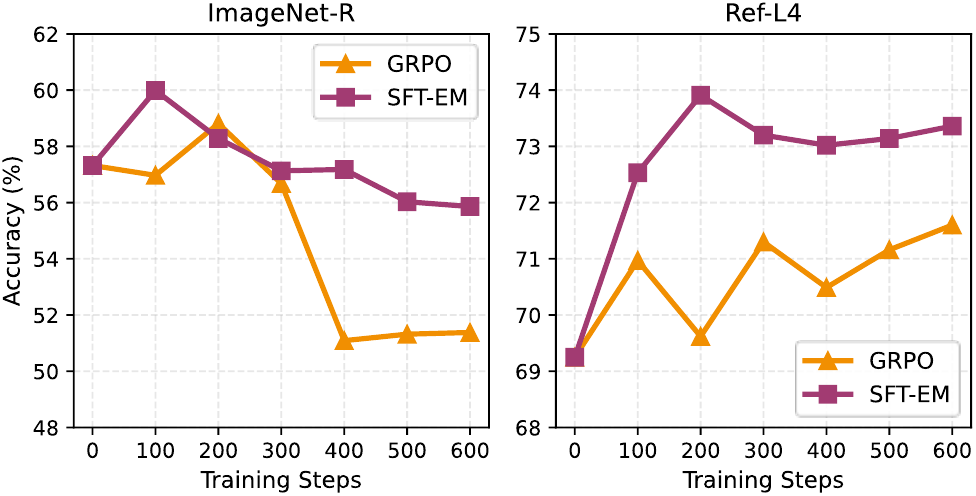}
   \caption{Performance curves of different post-training paradigms using Qwen2.5-VL-7B as the backbone.}
   \label{fig:steps}
\end{figure}

Existing research consistently reports that RL-based training exhibits higher training instability compared to SFT. Our results corroborate this observation. As shown in Table~\ref{tab:dcsft}, GRPO underperforms significantly relative to DC-SFT on ImageNet-R and even falls below standard SFT on Ref-L4 when using Qwen2.5-VL-7B as the backbone. To further investigate this instability, we analyze training dynamics by plotting performance curves over training steps in Figure~\ref{fig:steps}. While both GRPO and DC-SFT follow broadly similar trajectories, GRPO suffers from sharp performance drops during training. Specifically, it exhibits a pronounced decline between 300–400 steps on ImageNet-R (image classification) and another between 100–200 steps on Ref-L4 (visual grounding). In contrast, DC-SFT maintains consistently stable performance throughout training. These findings underscore the superior training stability of DC-SFT over RL methods such as GRPO.

\subsection{Efficiency Analysis} 

\begin{figure}[ht]
  \centering
  \includegraphics[width=1\linewidth]{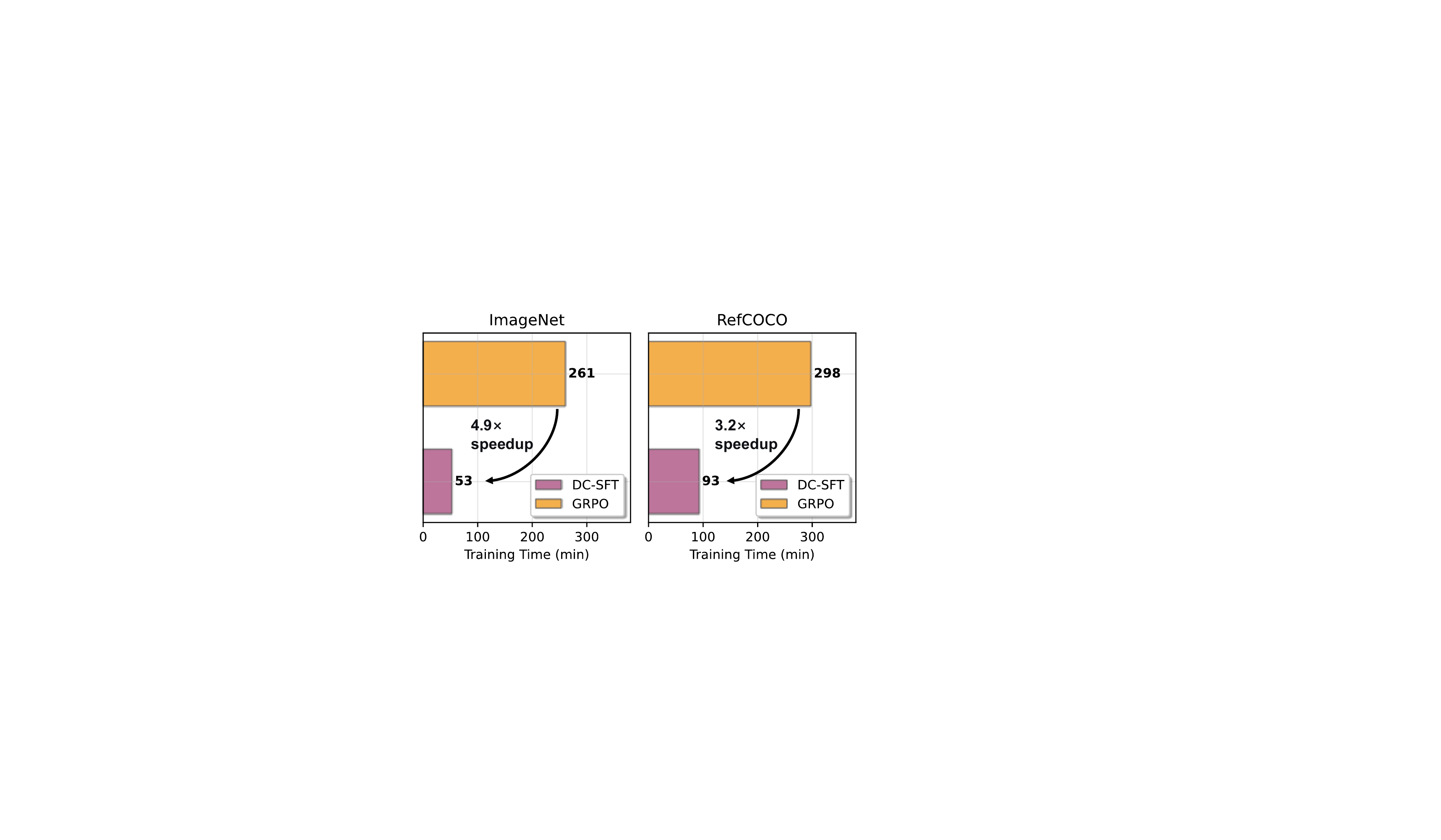}
   \caption{Training time comparison of Qwen2.5-VL-7B on ImageNet and RefCOCO.}
   \label{fig:efficiency}
\end{figure}

Although RL methods like GRPO achieve remarkable performance, they require time-consuming iterative sampling and model updates, which hinders their efficiency. In this subsection, we investigate whether the more direct DC-SFT approach can overcome this bottleneck. Figure~\ref{fig:efficiency} compares the training efficiency between GRPO and DC-SFT. The total time for DC-SFT includes both the process of constructing the difficulty-curated subset and the actual training time. Results demonstrate that DC-SFT achieves 4.9$\times$ higher efficiency on ImageNet and 3.2$\times$ higher efficiency on RefCOCO compared to GRPO, while delivering comparable or superior performance as shown in previous sections. This significant efficiency advantage, combined with its strong performance on both ID and OOD benchmarks, positions DC-SFT as a highly practical and scalable method for VLMs post-training.

\subsection{Scalability analysis}

\textbf{Evaluation on MiniCPM.} To further validate the general applicability of our method, we conduct additional experiments using MiniCPM-V-4~\cite{yao2024minicpm} as the foundational model. For this evaluation, we focus on the image classification task, maintaining experimental settings consistent with previous sections. As summarized in Table~\ref{tab:mini_dcsft}, both variants of our proposed DC-SFT approach consistently outperform standard SFT and GRPO in terms of OOD performance, corroborating our earlier findings. These results demonstrate that the effectiveness of DC-SFT extends across diverse VLMs.

\begin{table}[htbp]
\centering
\setlength{\tabcolsep}{1mm}
\caption{ID and OOD (gray background) performance (\%) of different post-training paradigms for MiniCPM-V-4.}
\label{tab:mini_dcsft}
\resizebox{\linewidth}{!}{
\begin{tabular}{cc>{\columncolor{lightgray}}c>{\columncolor{lightgray}}c}
\toprule
Paradigms & ImageNet & ImageNet-R & ImageNet-A \\
\midrule
SFT & \textbf{89.34} & 48.56 & 29.59 \\
GRPO & 83.22 (-6.12) & 56.70 (+8.14) & 30.35 (+0.76) \\
SFT-M & \underline{89.30} (-0.04) & \underline{59.72} (+11.16) & \underline{31.54} (+2.95) \\
SFT-EM & 84.22 (-5.12) & \textbf{60.91} (+12.35) & \textbf{31.67} (+3.08) \\
\bottomrule
\end{tabular}
}
\end{table}

\textbf{Evaluation on Larger Training Set.} We further investigate the scalability of our approach under increased data conditions. To this end, we construct a larger subset of ImageNet-1K containing 200 classes with 500 samples each, yielding 100k training examples. ID evaluation is conducted on the corresponding 200-class test split. As shown in Table~\ref{tab:100k}, DC-SFT (SFT-M) consistently outperforms standard SFT on OOD benchmarks and achieves performance on par with or surpassing GRPO across both model scales. These results reinforce the conclusions drawn from prior experiments and confirm that DC-SFT remains effective as training data increases.

\begin{table}[h]
\centering
\setlength{\tabcolsep}{1mm}
\caption{ID and OOD (gray background) performance (\%) of different post-training paradigms with 100k training samples.}
\label{tab:100k}
\resizebox{\linewidth}{!}{
\begin{tabular}{ccc>{\columncolor{lightgray}}c>{\columncolor{lightgray}}c}
\toprule
Model &Method &ImageNet & ImageNet-R & ImageNet-A \\
\midrule
\multirow{3}{*}{\makecell[c]{Qwen2.5-\\VL-3B}} &SFT &\textbf{93.03} & 59.33 & 46.00 \\
& GRPO & \underline{91.64} (-1.39) & \underline{65.79} (+6.46) & \underline{48.33} (+2.33) \\
& SFT-M & 90.69 (-2.34) & \textbf{70.09} (+10.76) & \textbf{48.92} (+2.92) \\
\midrule
\multirow{3}{*}{\makecell[c]{Qwen2.5-\\VL-7B}} &SFT &\textbf{94.26} & 63.16 & 52.44\\
& GRPO & \underline{93.16} (-1.10) & \underline{66.19} (+3.03) & \underline{53.86} (+1.42) \\
& SFT-M & 92.16 (-2.10) & \textbf{66.94} (+3.78) & \textbf{54.20} (+1.76) \\
\bottomrule
\end{tabular}
}
\end{table}

\textbf{Evaluation with Full-Parameter Fine-Tuning.} While the preceding experiments primarily adopt LoRA-based efficient fine-tuning, we further examine whether DC-SFT retains its effectiveness under full-parameter fine-tuning. Results are presented in Table~\ref{tab:full}. Consistent with our LoRA-based findings, DC-SFT (SFT-M) outperforms standard SFT on OOD tasks and delivers competitive or superior performance relative to GRPO. This suggests that the benefits of DC-SFT can generalize to full-parameter adaptation scenarios.

\begin{table}[h]
\centering
\setlength{\tabcolsep}{1mm}
\caption{ID and OOD (gray background) performance (\%) of different post-training paradigms using full-parameter training.}
\label{tab:full}
\resizebox{\linewidth}{!}{
\begin{tabular}{ccc>{\columncolor{lightgray}}c>{\columncolor{lightgray}}c}
\toprule
Model &Method &ImageNet & ImageNet-R & ImageNet-A \\
\midrule
\multirow{3}{*}{\makecell[c]{Qwen2.5-\\VL-3B}} &SFT &\textbf{95.52} & 34.99 & 27.13 \\
& GRPO & 89.74 (-5.78) & \textbf{50.61} (+15.62) & \underline{27.86} (+0.73) \\
& SFT-M & \underline{93.60} (-1.92) & \underline{45.03} (+10.04) & \textbf{30.78} (+3.65) \\
\midrule
\multirow{3}{*}{\makecell[c]{Qwen2.5-\\VL-7B}} &SFT &\textbf{96.04} & 28.38 & 26.63\\
& GRPO & 87.14 (-8.90) & \underline{46.68} (+18.30) & \underline{34.10} (+7.47) \\
& SFT-M & \underline{94.58} (-1.46) & \textbf{46.84} (+18.46) & \textbf{34.43} (+7.80) \\
\bottomrule
\end{tabular}
}
\end{table}

\section{Analysis and Discussion}

\subsection{Hard Data's Impact on SFT Generalization}

\begin{figure}[ht]
  \centering
  \includegraphics[width=1\linewidth]{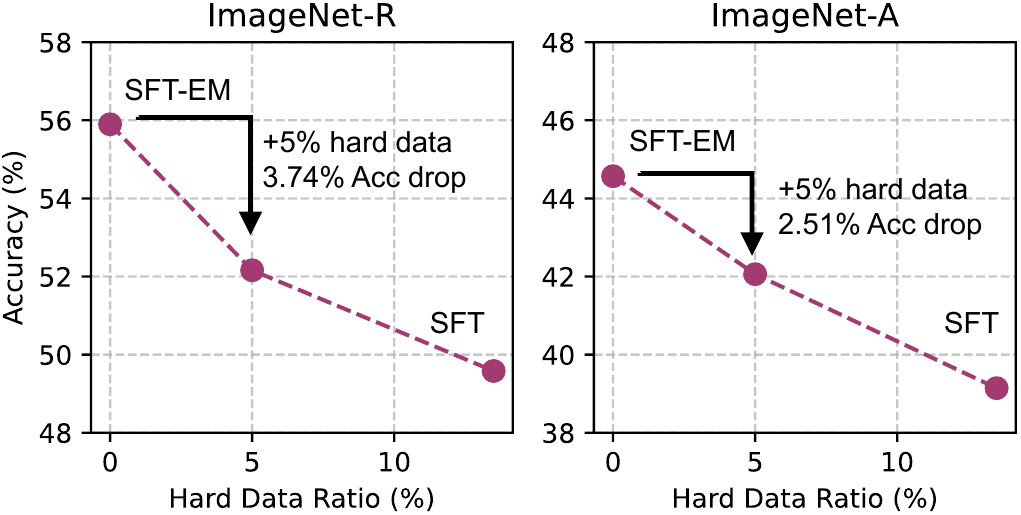}
   \caption{The impact of hard data ratio on OOD performance.}
   \label{fig:hard2acc}
\end{figure}

Our findings indicate that hard data exert a substantial influence on model generalization, despite constituting only a minor portion of the overall training set. To quantify this effect, we perform a controlled ablation by incrementally adding hard samples into the SFT process. Figure~\ref{fig:hard2acc} illustrates how varying proportions of hard data affect OOD performance, using Qwen2.5-VL-7B as the base model. We observe that standard SFT, which incorporates only 13.5\% hard data, leads to a notable drop in OOD accuracy compared to SFT-EM, which uses only easy and medium-difficulty samples. Remarkably, even a modest 5\% inclusion of hard samples causes significant degradation, reducing accuracy by 3.74\% on ImageNet-R and 2.51\% on ImageNet-A relative to SFT-EM. This confirms that the detrimental impact of hard data is not a threshold effect, but emerges clearly even at low mixing ratios.

\begin{figure}[ht]
  \centering
  \includegraphics[width=1\linewidth]{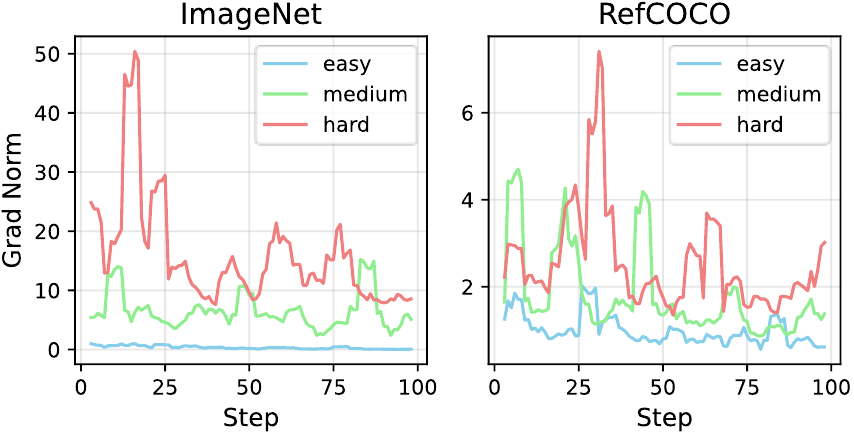}
   \caption{Gradient norms observed during SFT training on data subsets of varying difficulty.}
   \label{fig:grad_norm}
\end{figure}

To investigate why a small number of hard samples can severely impair generalization, we analyze gradient dynamics across data subsets of varying difficulty during SFT. Figure~\ref{fig:grad_norm} plots the per-step gradient norms for Qwen2.5-VL-7B trained on easy, medium, and hard subsets from both ImageNet and RefCOCO. The results reveal a consistent stratification: hard data produce significantly larger gradient norms throughout training compared to easy or medium-difficulty data. This indicates that hard samples dominate the optimization trajectory by inducing more aggressive parameter updates during SFT. This may encourage overfitting to noisy or ambiguous patterns and ultimately undermine robustness on OOD benchmarks.

\begin{table*}[htbp]
\centering
\caption{Reasoning performance (\%) of models built using different post-training paradigms. We only assess multiple-choice questions to ensure an objective evaluation. \textbf{Bolded} indicates the best.}
\label{tab:tts}
\begin{tabular}{ccccccccc}
\toprule
Model &Paradigms & MMK12 & MMMU & WeMath & MathVerse & MathVista & MathVision & Average\\
\midrule
\multirow{4}{*}{\makecell[c]{Qwen2.5-\\VL-7B}} 
& SFT & 42.20 & 47.45 & 54.14 & 49.54 & 68.33 &29.50 &48.53 \\
& SFT-RS & 41.80 & 47.45 & 52.41 & 48.72 & 65.19 &29.83 &47.57 \\ 
& GRPO & 42.10 & 48.56 & 52.82 & 48.67 & 66.48 &29.96 &48.10 \\
& SFT-M & \textbf{43.40} & \textbf{50.11} & \textbf{54.89} & \textbf{51.51} & \textbf{68.89} &\textbf{31.92} & \textbf{50.12} \\
\midrule
\multirow{4}{*}{\makecell[c]{Qwen2.5-\\VL-7B}}
& SFT & 49.20 & 51.67 & 60.23 & 58.39 & 71.67 &32.11 &53.88 \\ 
& SFT-RS & 48.60 & 50.33 & 59.20 & 55.87 & 73.52 &32.25 &53.20 \\
& GRPO & 49.05 & 51.89 & 57.99 & 57.16 & 69.81 &32.83 &53.12 \\
& SFT-M & \textbf{50.70} & \textbf{52.56} & \textbf{62.30} & \textbf{59.31} & \textbf{74.07} &\textbf{32.96} &\textbf{55.32} \\
\bottomrule
\end{tabular}
\end{table*}

\subsection{Test-Time Scaling Analysis}
Recent research shows that RL can effectively benefit test-time scaling by encouraging models to perform complex reasoning. Specifically, training on mathematical or coding data enables RL-trained models to exhibit strong reasoning performance on OOD tasks. Moreover, recent studies indicate that SFT combined with model distillation can enhance reasoning abilities, sometimes even outperforming RL for relatively small models~\cite{guo2025deepseek}. In this study, we explore whether DC-SFT can also contribute to test-time scaling.

\textbf{Training Data and Methods.}
We source our training data from the MMK12 dataset~\cite{meng2025mm}. A large teacher model (GLM-4.5) is leveraged to generate high-quality reasoning responses for SFT, producing 6,800 verified correct question–response pairs. Of these, 400 pairs are reserved for model initialization, and the remaining 6,400 are used in subsequent post-training. We compare the SFT-M variant of our proposed DC-SFT method against standard SFT, GRPO, and a Rejection Sampling (RS) baseline, where the model is trained only on its own self-generated correct responses. For GRPO, we apply a KL divergence penalty of 0.001 to regularize the model, following the best practice established by the open-source library ms-swift~\cite{zhao2024swiftascalablelightweightinfrastructure}. All models are trained for 400 steps, with other hyperparameters matching those described in Section~\ref{sec:settings}.

\textbf{Benchmarks and Evaluation Settings.} We evaluate our models on several major datasets: MMK12~\cite{meng2025mm}, MMMU (val)~\cite{yue2024mmmu}, WeMath~\cite{qiao2025we}, MathVerse (testmini)~\cite{zhang2024mathverse}, MathVista (testmini)~\cite{lu2023mathvista}, and MathVision (test)~\cite{wang2024measuring}. MMK12’s test split evaluates the ability to solve fundamental, multidisciplinary K–12 problems across Mathematics, Physics, Chemistry, and Biology. MMMU serves as a multimodal general knowledge benchmark. The remaining four datasets (i.e., WeMath, MathVerse, MathVista, and MathVision) are representative multimodal mathematics benchmarks designed to assess knowledge mastery and generalization.

To ensure objective evaluation, we assess only multiple-choice questions. In particular, all non-multiple-choice questions in MMMU are converted into a multiple-choice format, following conventions in~\cite{bai2025qwen2}. During evaluation, we employ greedy decoding with a temperature of 0, a repetition penalty of 1.1, and a maximum token limit of 4096.

\textbf{Performance Results.}
The results are presented in Table~\ref{tab:tts}. We observe that when using question–response pairs distilled from a large teacher model, SFT yields stronger reasoning performance than GRPO, consistent with existing findings~\cite{guo2025deepseek}. More importantly, SFT-M consistently achieves the highest performance across all reasoning benchmarks, surpassing standard SFT. For the Qwen2.5-VL-3B model, SFT-M outperforms all other paradigms, achieving an average score of 50.12\%, which represents a 1.59\% improvement over standard SFT. Similarly, for the Qwen2.5-VL-7B model, SFT-M attains the best results and reaches the highest average score of 55.32\%, exceeding standard SFT by 1.44\%. These results confirm that the advantages of DC-SFT extend beyond OOD generalization, offering a more efficient and stable pathway for enhancing downstream reasoning capabilities.

\section{Conclusion}

In this work, we present a data-centric explanation for the generalization gap between RL and SFT in VLM post-training. We find this gap stems from an implicit data filtering mechanism in RL, which naturally focuses updates on medium-difficulty samples. In contrast, standard SFT's generalization is degraded by a small fraction of hard samples that dominate the optimization. Based on this insight, we propose DC-SFT, a simple yet effective method that explicitly filters out hard data before training. Our experiments show that DC-SFT exceeds the generalization performance of strong RL baselines. Furthermore, DC-SFT achieves this with significantly greater training stability and computational efficiency. These findings demonstrate that DC-SFT offers a more efficient and stable path toward building VLMs with strong generalization ability.

\textbf{Limitations.} 
We acknowledge several limitations in this work. First, our validation is restricted to a relatively narrow set of model architectures, primarily Qwen2.5-VL and MiniCPM-V-4. Second, the scale of our experiments is limited to models with up to 7B parameters. Third, while we include some full-parameter fine-tuning results, the majority of our experiments employ LoRA fine-tuning.

{
    \small
    \bibliographystyle{ieeenat_fullname}
    \bibliography{main}
}


\end{document}